%% file: PaperForArxiv.tex
\documentclass[10pt,twocolumn,letterpaper]{article}

\usepackage[pagenumbers]{wacv} %

\usepackage{graphicx}
\usepackage{amsmath}
\usepackage{amssymb}
\usepackage{booktabs}
\usepackage{caption} %
\usepackage{adjustbox} %
\usepackage{svg}
\usepackage{dblfloatfix}
\usepackage{siunitx} %
\usepackage{wrapfig}
\usepackage{amssymb}
\usepackage{graphicx}

\input{0-imports}

\input{0-macros}

\input{0-math}

\input{0-space-saving}

\usepackage[accsupp]{axessibility}

\usepackage[pagebackref,breaklinks,colorlinks]{hyperref}

\usepackage[capitalize]{cleveref}
\crefname{section}{Sec.}{Secs.}
\Crefname{section}{Section}{Sections}
\Crefname{table}{Table}{Tables}
\crefname{table}{Tab.}{Tabs.}

\def\wacvPaperID{1696} %
\def\confName{WACV}
\def\confYear{2025}

\begin{document}

\title{Patch Ranking: Token Pruning as Ranking Prediction for Efficient CLIP}

\author{Cheng-En Wu\thanks{Equal Contribution} ~~~~Jinhong Lin\textsuperscript{*} ~~~~Yu Hen Hu ~~~~Pedro Morgado  \\
University of Wisconsin–Madison \\
{\tt\small \{cwu356, jlin398, yhhu, pmorgado\}@wisc.edu}}

\maketitle
\input{content/0-abstract}
\input{content/1-intro}
\input{content/2-related-work}

\input{content/3-method}

\input{content/4-experiments}
\input{content/5-conclusion}

\input{content/6-acknowledgement}

{\small
\bibliographystyle{ieee_fullname}
\bibliography{egbib}
}

\clearpage
\appendix 
\input{content/7-supplementary}

\end{document}

%% file: 0-imports.tex
\usepackage[utf8]{inputenc} %
\usepackage[T1]{fontenc}    %
\usepackage{url}            %
\usepackage{nicefrac}       %
\usepackage{microtype}      %
\usepackage{xcolor}         %
\definecolor{lightgray}{rgb}{0.9, 0.9, 0.9}
\definecolor{tabhighlight}{HTML}{e5e5e5}
\usepackage{amsmath}
\usepackage{amsfonts}
\usepackage{bm}

\usepackage{booktabs}   %
\usepackage{tabularx}   %
\usepackage{colortbl}   %
\usepackage{multirow}
\usepackage{makecell}

\newcommand{\mc}[3]{\multicolumn{#1}{#2}{#3}}
\newcommand{\mr}[2]{\multirow{#1}{*}{#2}}

\usepackage{wrapfig}
\usepackage{mwe} %

\usepackage[font=normal,labelfont=bf]{caption}
\usepackage[font=normal]{subcaption}

\usepackage[normalem]{ulem} %
\usepackage{array}

\usepackage{xparse}
\usepackage{pifont}

\usepackage{ifthen}

\definecolor{cvprblue}{rgb}{0.21,0.49,0.74}
\usepackage[pagebackref,breaklinks,colorlinks,citecolor=cvprblue]{hyperref}

\usepackage[capitalize]{cleveref}
\crefformat{section}{Section~#2#1#3}
\crefformat{subsection}{Section~#2#1#3}
\crefformat{subsubsection}{Section~#2#1#3}
\crefformat{figure}{Fig.~#2#1#3}
\crefformat{equation}{Eq.~#2#1#3}
\crefformat{table}{Table~#2#1#3}
\crefformat{algorithm}{Alg.~#2#1#3}
\usepackage{algpseudocode}
\usepackage[linesnumbered,ruled,vlined]{algorithm2e}

\SetCommentSty{mycommfont}
\SetKwInput{KwInput}{Input}

\usepackage{enumitem}

\usepackage{xfp}

%% file: 0-macros.tex
\newcommand{\cmark}{\ding{51}}%

\newlength\savewidth
\newlength\thinwidth

\definecolor{Gray}{gray}{0.93}
\newcolumntype{a}{>{\columncolor{Gray}}c}

\usepackage{xspace} %
\makeatletter
\DeclareRobustCommand\onedot{\futurelet\@let@token\@onedot}
\def\@onedot{\ifx\@let@token.\else.\null\fi\xspace}
 
\def\ie{\emph{i.e}\onedot} 
 
 \def\vs{\emph{vs}\onedot}

\makeatother

%% file: 0-math.tex
\def\1{\bm{1}}

\def\vs{{\bm{s}}}

\def\mW{{\bm{W}}}
\def\mX{{\bm{X}}}

\def\mZ{{\bm{Z}}}

\DeclareMathAlphabet{\mathsfit}{\encodingdefault}{\sfdefault}{m}{sl}
\SetMathAlphabet{\mathsfit}{bold}{\encodingdefault}{\sfdefault}{bx}{n}

\def\gL{{\mathcal{L}}}

\def\gR{{\mathcal{R}}}

\def\gT{{\mathcal{T}}}

%% file: 0-space-saving.tex
\usepackage{enumitem}
\setlist{itemsep=1pt, topsep=3pt}

\renewcommand{\paragraph}[1]{\noindent {\bf #1}}

%% file: content/0-abstract.tex
\begin{abstract}
Contrastive image-text pre-trained models such as CLIP have shown remarkable adaptability to downstream tasks. However, they face challenges due to the high computational requirements of the Vision Transformer (ViT) backbone. Current strategies to boost ViT efficiency focus on pruning patch tokens but fall short in addressing the multimodal nature of CLIP and identifying the optimal subset of tokens for maximum performance. To address this, we propose greedy search methods to establish a ``Golden Ranking'' and introduce a lightweight predictor specifically trained to approximate this Ranking. To compensate for any performance degradation resulting from token pruning, we incorporate learnable visual tokens that aid in restoring and potentially enhancing the model's performance. Our work presents a comprehensive and systematic investigation of pruning tokens within the ViT backbone of CLIP models. 
Through our framework, we successfully reduced 40\% of patch tokens in CLIP's ViT while only suffering a minimal average accuracy loss of 0.3\% across seven datasets. Our study lays the groundwork for building more computationally efficient multimodal models without sacrificing their performance, addressing a key challenge in the application of advanced vision-language models.\footnote{Project Page: \url{https://github.com/CEWu/PatchRanking}}
\end{abstract}

%% file: content/1-intro.tex
\section{\label{sec:intro}Introduction}

Contrastive Language-Image Pretraining (CLIP)~\cite{radford2021CLIP} has emerged as a paradigm shift in the field of visual recognition, demonstrating remarkable transferability across a wide array of downstream tasks through zero-shot inference. By training models to align representations of images with text descriptions at scale (400 million text-image pairs in~\cite{radford2021CLIP}), language-image pertaining enables zero-shot (or open-dictionary) recognition by matching the learned visual embeddings to class embeddings constructed from hand-crafted text prompts such as ``a photo of a [class]''. Although CLIP's success is undeniable, commonly used CLIP backbones like the Vision Transformer (ViT)~\cite{vits} can be computationally intensive during inference. The complexity of this process increases quadratically with the length of the tokens in the self-attention layer, posing significant challenges for practical deployment.

\begin{figure}[t]
  \begin{center}
    \includegraphics[width=0.50\textwidth]{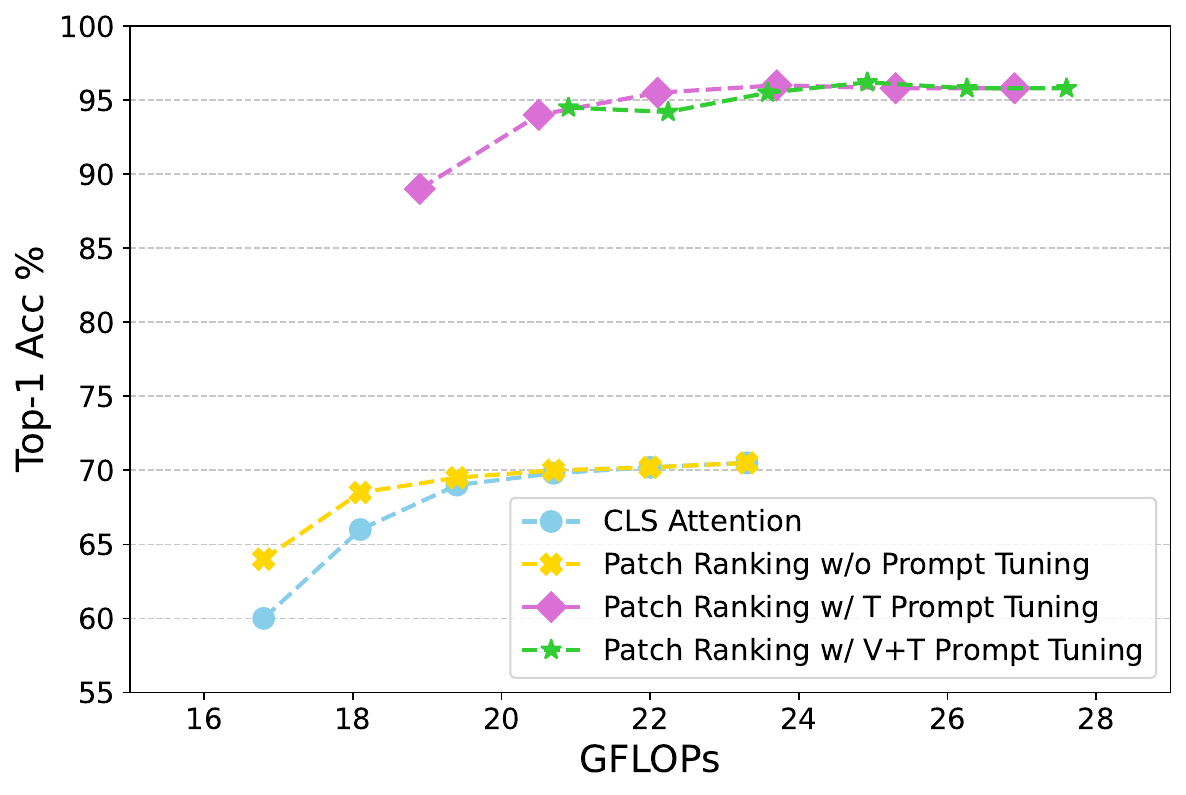}
  \end{center}
    \vspace{-1.4ex}
  \caption{\label{fig:performance_rate_change} Accuracy vs. complexity for various token pruning strategies in pre-trained CLIP models is evaluated on Caltech101. Six points represent models with token keep-rates from 100\% to 50\%. The CLS Attention method prunes image patches by measuring similarity between CLS tokens and others in the 4th layer of CLIP's ViT. Patch Ranking, using our Preservation-based ranking strategy, outperforms the traditional CLS method. Patch Ranking w/ T Prompt Tuning and Patch Ranking w/ V+T Prompt Tuning extend this by adding learnable tokens to the Text Encoder or both the ViT and Text Encoder, fine-tuned with 16 shots per class. Prompt-tuning boosts performance, and tuning both prompts (green line) shows no significant degradation up to a 50\% keep rate.}
\end{figure}

One of the most direct and effective strategies to alleviate this computational burden is token pruning, a method that has received considerable attention recently~\cite{wei2023joint, Chen2023DiffRate, long2023beyond, yin2022avit, meng2022adavit, tang2022patchslim, kong2022spvit, fayyaz2022adpt, xu2022evo_vit, Liang2022EViTEV, Rao2021DynamicViTEV}. Current studies in this domain focus mainly on the design of various metrics to assess token importance to eliminate those deemed redundant. However, these approaches~\cite{long2023beyond, Liang2022EViTEV, xu2022evo_vit} typically rely on the attention between the CLS and patch tokens. 
However, it is unclear what these attention weights capture, particularly in the early stages of the model. 

To address this issue, we propose a framework in which pruning is guided by well-defined and interpretable scoring functions. Our method follows three phases: Phase I ranks each token based on three scoring functions that measure the usefulness of each token for either (1) optimal classification, (2) maximum confidence in the model's prediction, or (3) minimal impact on the model's output representation. These scoring functions establish the ``Golden Ranking'' of tokens. While useful, the Golden Ranking does not necessarily speed up inference, as the full sequence must be evaluated to compute it. Thus, in Phase II, we introduce a lightweight predictor, trained to closely approximate the Golden Ranking, and thus to determine which tokens to prune during inference. Finally, Phase III addresses potential performance drops due to token pruning by tuning the model to operate on pruned sequences. Given our focus on resource efficiency (both during inference and training), we demonstrate that prompt tuning techniques, \ie, by integrating both learnable visual and text tokens, effectively recover the model performance with minimal training budget requirements. Through systematic experiments, on a large number of datasets, we validated the effectiveness of our method. Our results demonstrate that the proposed framework significantly reduces computational complexity without compromising the model's classification accuracy.
As a result, this work presents a compelling solution that balances model efficiency and effectiveness, opening new avenues for the practical application of CLIP models in various real-world settings. Our contributions are summarized as follows:

\begin{itemize}
\item We propose the ``Golden Ranking'' which ranks patch tokens in CLIP models based on their usefulness to the model's predictive capabilities. 

\item We introduce a lightweight predictor, specifically trained to approximate the Golden Ranking, to guide the token pruning process in CLIP models during inference.

\item We demonstrate that the integration of learnable tokens into the CLIP model compensates for the inevitable performance degradation due to token pruning, effectively enhancing the model's accuracy post-pruning.
\end{itemize}

%% file: content/2-related-work.tex
\begin{figure*}[t]
    \vspace{5pt}
      \begin{center}
        \includegraphics[width=0.95\textwidth]{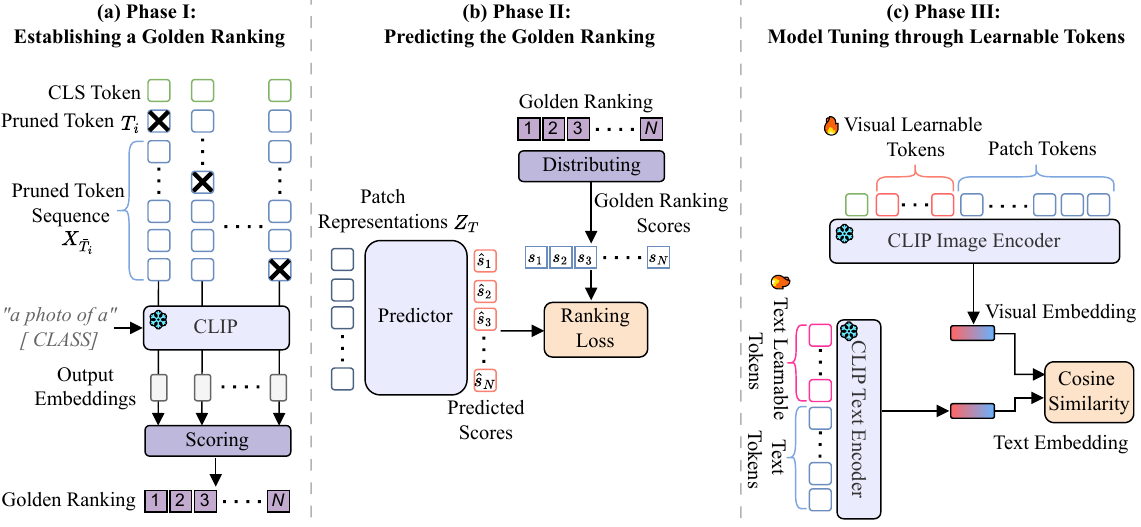}
      \end{center}
    \caption{\label{fig:framework_diagram}This diagram presents an overview of our pruning framework for patch tokens in CLIP's ViT. The framework comprises three main phases: \textbf{(a) Phase I: Establishing a Golden Ranking}, which involves assigning scores to each token based on their importance, as discussed in Section~\ref{sec:golden_ranking}; \textbf{(b) Phase II: Predicting the Golden Rankin}, which focuses on training a predictor to approximate the Golden Ranking, as elaborated in Section~\ref{sec:predictor}; and \textbf{(c) Phase III: Model Tuning through Learnable Tokens}, a process where additional visual learnable tokens are added to mitigate accuracy loss resulting from the removal of patch tokens, detailed in Section~\ref{sec:prompt_tuning}.}
    \end{figure*}

\section{\label{sec:related-work}Related Work}

\paragraph{Token Pruning.}
The efficiency of ViT is crucial, especially because their attention mechanisms require a lot of computational resources. A direct and intuitive approach to enhance ViT efficiency is the reduction of patch tokens, especially considering that some of these tokens may be redundant. Various studies have proposed approaches to evaluate and prune less informative tokens. These approaches broadly fall into two categories. The first category leverages the weights of patch tokens as attended by the CLS token, effectively identifying tokens with lesser contributions to the overall model prediction~\cite{long2023beyond, Liang2022EViTEV, xu2022evo_vit}. The second category involves the integration of additional learnable modules within the ViT architecture~\cite{meng2022adavit, Rao2021DynamicViTEV, pan2021ia_red}. However, neither of these methods conclusively demonstrates that the tokens pruned are indeed the optimal subset, which would allow the model to achieve the highest possible accuracy for a given pruning ratio. In our token pruning approach, we introduce ``Golden Ranking'' — an optimal ranking of tokens determined by our proposed metrics. This ranking acts as a ground truth for a newly introduced lightweight predictor within ViTs. By learning from the Golden Ranking, the predictor can efficiently discern which tokens to prune, striking a balance between maintaining model accuracy and enhancing computational efficiency.

\paragraph{Prompt Tuning.}
Prompt tuning is a new paradigm for adapting pre-trained models for various tasks and domains. It includes text prompt tuning in natural language processing (NLP), which has progressed from using handcrafted prompts in models like GPT-3~\cite{brown2020lgpt3} to learnable prompts for better unimodal task performance~\cite{liu2022ptuning, Lester2021nlppt}.  In computer vision, VPT~\cite{jia2022vpt} introduces visual learnable tokens for ViT to enhance the transfer performance of downstream tasks. Recently, prompt tuning has been extended to vision-language (V-L) pre-trained models. For example, CoOp~\cite{Zhou2022CoOp} adapts CLIP for downstream tasks by optimizing learnable text prompts. Co-CoOp~\cite{zhou2022CoCoOp} builds on this by introducing a meta-net that integrates image-conditional context with text prompts.
Despite significant progress in prompt learning for VL models, most methods primarily target the text encoder, neglecting the image encoder's adaptation, which can result in less optimal visual features. To overcome this, MaPLe~\cite{khattak2023maple} aims to simultaneously learn both vision and language prompts.

\paragraph{Learning to Rank.}
Learning-to-rank has gained prominence in machine learning, primarily utilizing a score-and-sort strategy to solve ranking problems. The main aim of these approaches is to create scoring functions that determine the relevance of individual items, which are then used to derive a ranking order. Learning-to-rank has found widespread application in various fields, notably in information retrieval~\cite{liu2009retrieval} and recommendation systems~\cite{karatzoglou2013recommender}. In this paper, we introduce an innovative application within this field. To our knowledge, it is the first work to adapt the learning-to-rank approach for pruning patch tokens in ViT. We develop a lightweight module that evaluates and ranks the importance of patch tokens. This ranking then guides the pruning process, removing tokens of lesser significance. Such an approach significantly boosts the efficiency of ViT models.

%% file: content/3-method.tex
\section{Method\label{sec:method}}
Unlike existing methods that predominantly leverage CLS attention for token ranking --- a technique not ideally suited for CLIP due to its multi-modal embedding architecture --- we introduce a novel token pruning strategy, illustrated in \cref{fig:framework_diagram}.
The proposed framework unfolds into three distinct stages: (a) Searching for the ``Golden Ranking'' of patch tokens; (b) Learning to rank patch tokens by training a predictor to approximate the Golden Ranking; and (c) Compensating for the potential performance degradation incurred after removal of uninformative tokens. 
These three stages, elaborated in the subsequent sections, collectively form the foundation of our approach and enable the deployment of CLIP models with significant speed enhancements and minimal performance degradation.

\subsection{\label{sec:golden_ranking}Phase I:~Establishing a Golden Ranking}
Token pruning can be conceptualized as a problem of identifying the optimal subset of tokens that maximizes the model's accuracy. While a brute force search could in principle identify this optimal subset, it is impractically time-consuming. Consequently, most existing methods resort to heuristic approaches, such as utilizing CLS attention weights, to determine suitable subsets for pruning. However, these methods often fall short of truly approximating the optimal pruning subset. 
To bridge this gap, we introduce three scoring metrics designed to rank patch tokens based on their impact on CLIP's predictions, forming what we term the `Golden Ranking`.

\cref{fig:framework_diagram} (a) illustrates how the Golden Ranking is obtained. We start with a set of class prompts and an image that has been subdivided and encoded into a token sequence $\mX$ of length $N$. Normally, CLIP's ViT encoder $f_i$ would process this token sequence $\mZ, \mZ^{cls} = f_i(\mX)$, to obtain a set of visual embeddings $\mZ$ and a CLS token embedding $\mZ_{cls}$.
However, to determine the Golden Ranking, we apply the CLIP model to a series of pruned token sequences $\mX_{\bar{\gT_i}}$, each of which is obtained by removing a small set of tokens $\gT_i$ from $\mX$. The resulting visual embeddings $\mZ_{\bar{\gT_i}}, \mZ_{\bar{\gT_i}}^{cls} = f_i(\mX_{\bar{\gT_i}})$ can then used to determine the importance of each token in $\gT_i$.  To measure this score, denoted $s(\gT_i)$, we experimented with three distinct metrics: 
\begin{description}[leftmargin=3.2ex]
    \item[(1) Label-Driven Ranking Score] The pruned tokens $\gT_i$ are scored based on CLIP's zero-shot posterior probability of assigning the pruned sequence $\mX_{\bar{\gT_i}}$ to the ground-truth label $y_\text{gt}$
    \begin{equation}
        s(\gT_i) = P(y_{\text{gt}} | \mX_{\bar{\gT_i}})
    \end{equation}
    A high score of $s(\gT_i)$ suggests that the removed tokens are not required for accurate classification, and some tokens might even be misleading, resulting in more accurate classification after removal.
    
    \item[(2) Maximum Confidence Score] Label-Driven Ranking Score requires prior knowledge of the ground-truth class. To avoid this assumption, we can alternatively assess the pruned tokens $\gT_i$ based on the maximum confidence across all classes. 
    \begin{equation}
        s(\gT_i) = \max_{y} P(y | \mX_{\bar{\gT_i}})
    \end{equation}
    A high score of $s(\gT_i)$ indicates that the removal of tokens in $\gT_i$ did not reduce (or even increase) the model's overall confidence in its prediction.

    \item[(3) Feature Preservation Score] Finally, instead of searching for tokens that optimize classification performance, which makes the golden ranking task-specific, feature preservation seeks to identify the tokens that, when removed, do not alter the image representation, as expressed by the CLS token embedding. This score is quantified using cosine similarity:
    \begin{equation}
        s(\gT_i) = \frac{\mZ^{cls} \cdot \mZ_{\bar{\gT_i}}^{cls}}{\|\mZ^{cls}\| \|\mZ_{\bar{\gT_i}}^{cls}\|}
    \end{equation}
    where $\mZ^{cls}$ denotes the CLS embedding obtained from the full sequence $\mX$ and $\mZ_{\bar{\gT_i}}^{cls}$ denotes the embedding obtained with a pruned sequence. 
\end{description}

The metrics delineated above measure the importance of a set of tokens $\gT_i$.
To determine the importance of individual tokens, a straightforward approach is to prune one at a time, \ie, $\gT_i={t_i}$. However, this results in small changes in the model's output, making it challenging to discern the relative importance of different tokens. Instead, we remove a larger $r\times r$ block of tokens, resulting in more noticeable changes. As the removal block $\gT_i$ slides over the image, each token is removed and assessed multiple times, thus stabilizing the final average score of each token $t$ 
\begin{equation}
    s(t) = \frac{1}{|\gT_i|} \sum_{i: t\in\gT_i} s(\gT_i).
\end{equation}

The time complexity of estimating the golden ranking scores $s(t)$ using the sliding window approach is $O(L)$, where $L$ is the number of patches in the sequence. However, in practice, we can significantly reduce the time required to score all tokens, by creating a batch with all pruned sequences $\mX_{\bar{\gT_i}}$, and processing them simultaneously through the model.

\input{tables/golden_ranking}
\subsection{\label{sec:predictor}Phase II:~Predicting the Golden Ranking}
After establishing the Golden Ranking using one of the three metrics above, we train a lightweight predictor, $\hat{\vs}=h(\mZ;\Theta)\in\Re^N$, to efficiently approximate it, and thus identify the least useful tokens from their representations $\mZ$. Since token removal must occur early on to reduce the computational complexity, we deployed the predictor on internal representations $\mZ^{(i)}$ obtained at an early layer $i$.

\paragraph{Predictor Design.} 
The predictor architecture is a single Mix-MLP layer~\cite{tolstikhin2021mlp}, chosen for its ability to efficiently capture contextual dependencies among tokens. The Mix-MLP performs two main functions: token mixing and channel mixing, implemented using MLP layers. Given a sequence of patch representations $\mZ_\gT \in \gR^{|\gT|\times{}d}$, the channel and token mixing processes are computed as

\begin{align}
    \mZ_\gT^\text{channel} & = \textit{MLP}_\text{channel}(\mZ_\gT) + \mZ_\gT\\
    \mZ_\gT^\text{token} & = \textit{MLP}_\text{token}(\mZ_\gT^\text{channel}) + \mZ_\gT^\text{channel}\\
\end{align}
where $\textit{MLP}_\text{channel}$ and $\textit{MLP}_\text{token}$ operate on the channel and token dimensions, respectively. The final score $\hat{s}(\mathcal{T})\in\gR^{|\gT|}$ is obtained by average pooling of the token representations across the channel dimension, $\hat{s}(\mathcal{T}) = \text{Avg}_\text{token}(\mZ_\gT^\text{token})$.
Each MLP layer consists of a fully connected layer accompanied by layer normalization and a GELU activation function.

\paragraph{Loss Function.} The predictor $\hat{\vs}=h(\mZ^{(i)}; \Theta)$ is trained to regress normalized golden ranking scores $s_t=\frac{s(t)-\mu_s}{\sigma_s}$, where $\mu_s$ and $\sigma_s$ denote the mean and standard deviation of $s(t)$ among all patch tokens $t$. Specifically, we minimize
\begin{equation}
    \gL(\Theta) = -\sum_t \sigma(s_t) \log \sigma\left(\hat{s}_t\right)
\end{equation}
where $\sigma(\cdot)$ denotes the sigmoid function. Although this loss does not aim to directly predict the ranking between patches (which is too difficult to accomplish from early-stage representations), it encourages the predictor to assign the highest scores to tokens at the top of the ranking and the lowest scores to those at the bottom.

\paragraph{Token Removal.} During inference, a predetermined number of patch tokens is removed. Since the Golden Ranking predictor operates on CLIP's intermediate representations, the full sequence \(\gT\) is maintained until the predictor is applied. The tokens with the lowest predicted scores are then removed, and the compressed sequence is passed through the rest of the model. We experiment with different removal rates to determine the optimal trade-off between speed and accuracy. We also explore the impact of progressively removing tokens in a layer-wise manner, starting from the predictor layer. Our results show that progressive pruning is more effective, as it allows the model to remove tokens in a more controlled manner.

\subsection{\label{sec:prompt_tuning}Phase III:~Model Tuning through Learnable Tokens}
Although the predictor can identify the least useful tokens, removing them can still degrade performance, as the model is forced to operate outside of its training distribution.
A variety of fine-tuning methods could be employed to recover the lost performance. For example, one could simply fine-tune the CLIP model on the original 400M image-text pairs using pruned visual input sequences. Although likely to succeed, this approach would be computationally expensive and data-intensive. Instead, inspired by prompt-tuning strategies~\cite{Zhou2022CoOp, khattak2023maple}, we introduce a set of learnable tokens into the CLIP model and train them using a small dataset to compensate for the performance degradation. As demonstrated in~\cite{Zhou2022CoOp} and~\cite{khattak2023maple}, adjusting a model through learnable tokens is significantly more data and compute efficient than fine-tuning.

Similarly to CoOp~\cite{Zhou2022CoOp}, we augment the input sequence to the CLIP's text encoder, $\mW_{t}$, with a set of $b$ learnable tokens $\{P_t^i \in \gR^{d_t}\}^b_{i=1}$, where $d_t$ is the text embedding dimension. 
We also introduce additional visual tokens to the CLIP image encoder. Following~\cite{khattak2023maple}, the new visual tokens $\{P_v^i\}^b_{i=1}$ are obtained from the text learnable tokens $\{P_t^i\}^b_{i=1}$ through a linear transformation $P_v^i = MP_t^i$, where $M \in \gR^{d_v \times d_t}$ is a learnable projection matrix and $d_v$ is the visual embedding dimension. The new input sequences, $\mW_t^* = \{P_t^i\}^b_{i=1} \cup \mW_{t}$ and $\mW_v^* = \{P_v^i\}^b_{i=1} \cup \mW_v$, allow the model to dynamically adapt the text and visual representations to better align with each other, compensating for the representational shifts introduced by removing image patches.

%% file: tables/golden_ranking.tex
\setlength{\tabcolsep}{7pt}
\begin{table*}[]
\centering
\resizebox{0.90\textwidth}{!}{
\centering
\begin{tabular}{|c|c|c|cccc|c|cccc|}
\hline\vspace{-0.4ex}
\multirow{3}{*}{\bf Dataset} &
\multirow{3}{*}{\bf \thead{Golden Ranking\\Score}} &
\multirow{3}{*}{\bf \thead{0-Shot Acc\\No Pruning}} &
\mc{4}{c|}{\bf Golden Ranking} & \mr{3}{\bf \thead{Matching\\Rate@100}} & \mc{4}{c|}{\bf Predicted Ranking}\\
&&&\mc{4}{c|}{\bf Keep rate} & & \mc{4}{c|}{\bf Keep rate} \\
&&& 80 & 70 & 60 & 50 & & 80 & 70 & 60 & 50 \\ \hline
\multirow{3}{*}{Caltech101} 
& Label &  &\bf94.9&\bf94.6&\bf94.0&\bf92.4& 56.7  &91.6 &89.1 &86.4 &83.9 \\ 
& Confidence & 93.5  &91.6&90.9&90.4&89.1& 57.0 &91.5 &88.4 &85.3 &83.4\\ 
& Preservation &  &92.8 &92.4 & 92.2 &91.4 &\bf78.0 &\bf93.6&\bf93.2&\bf93.4&\bf91.0\\ 
\hline
\multirow{3}{*}{OxfordPets} 
& Label & &\bf93.9&\bf94.6&\bf94.4&\bf93.8&59.8&87.7&87.1&85.9&\bf84.5\\ 
& Confidence & 89.5 &88.1&88.2&88.0&87.2&60.4&86.8 &86.0 &84.6 &83.1 \\ 
& Preservation && 88.9 &88.4 & 88.0 &85.4 &\bf78.2&\bf88.6 &\bf88.4 &\bf88.1 &83.5 \\ 
\hline
\multirow{3}{*}{Flowers102} 
& Label &&\bf80.6&\bf82.3&\bf83.2&\bf80.8& 56.8&68.8 &66.7 &65.7 &63.3 \\ 
& Confidence &70.5 &70.6&70.6&70.3&68.5& 57.8&68.5 &67.3 &65.6 &62.5  \\ 
& Preservation &&70.8 &70.4 &69.1&66.9&\bf71.9&\bf69.6 &\bf68.8 &\bf67.9 &\bf63.9  \\ 
\hline
\multirow{3}{*}{Food101} 
& Label &&\bf95.9&\bf96.4&\bf96.2&\bf95.0&55.5&83.6 &80.6	&77.4	&72.1 \\ 
& Confidence &86.0  &86.1&86.0&85.7&84.7&55.5&83.2 &80.3 &76.9 &71.7 \\ 
& Preservation &&85.8&85.4&84.1&79.5&\bf71.9&\bf85.1 &\bf84.2 &\bf82.8 &\bf75.8 \\ 
\hline
\multirow{3}{*}{FGVCAircraft} 
& Label &&\bf43.5&\bf43.6&\bf44.2&\bf43.5&60.1 &17.6 &17.3 &17.1 &17.1\\ 
& Confidence & 23.4  &24.2&23.8&23.9&23.5&61.4&18.6 &18.7 &18.2 &17.9\\ 
& Preservation &&25.0&24.8&24.5&23.4&\bf89.3 &\bf23.2	&\bf23.2 &\bf22.9 &\bf22.4\\ 
\hline
\multirow{3}{*}{DTD} 
& Label &&\bf53.1&\bf54.0&\bf54.7&\bf55.1&55.7 &44.6 &\bf44.6 &\bf45.0 &\bf43.3 \\ 
& Confidence &45.1  &44.3&44.1&44.4&44.3&56.1&44.9 &44.6 &44.7 &43.0 \\ 
& Preservation &&44.1&44.0&43.9&43.3&\bf72.3&\bf45.0 &44.2 &43.7 &43.0 \\ 
\hline
\multirow{3}{*}{UCF101} 
& Label &&\bf79.0&\bf79.1&\bf78.7&\bf77.1&55.9&65.2 &63.6 &60.0 &56.4  \\ 
& Confidence &67.0 &66.2&65.0&64.6&63.5&55.9&64.7 &62.3 &58.7 &54.6  \\ 
& Preservation &&66.6&67.4&67.0&65.2&\bf79.7&\bf66.6 &\bf65.9 &\bf65.2 &\bf60.9 \\ 
\hline
\end{tabular}}
\caption{Pruning effectiveness leveraging either the ground-truth or predicted golden ranking using three scoring functions: Label, Confidence, and Preservation. ``Matching Rate'' measures the agreement of the top 100 tokens in the ground-truth and predicted rankings. In all cases, token pruning was applied at the 4th layer of CLIP's ViT, and classification was conducted without any additional model tuning.}
\label{tbl:golden_ranking}
\end{table*}

%% file: content/4-experiments.tex
\begin{figure*}[h]
  \begin{center}
    \includegraphics[width=1.0\textwidth]{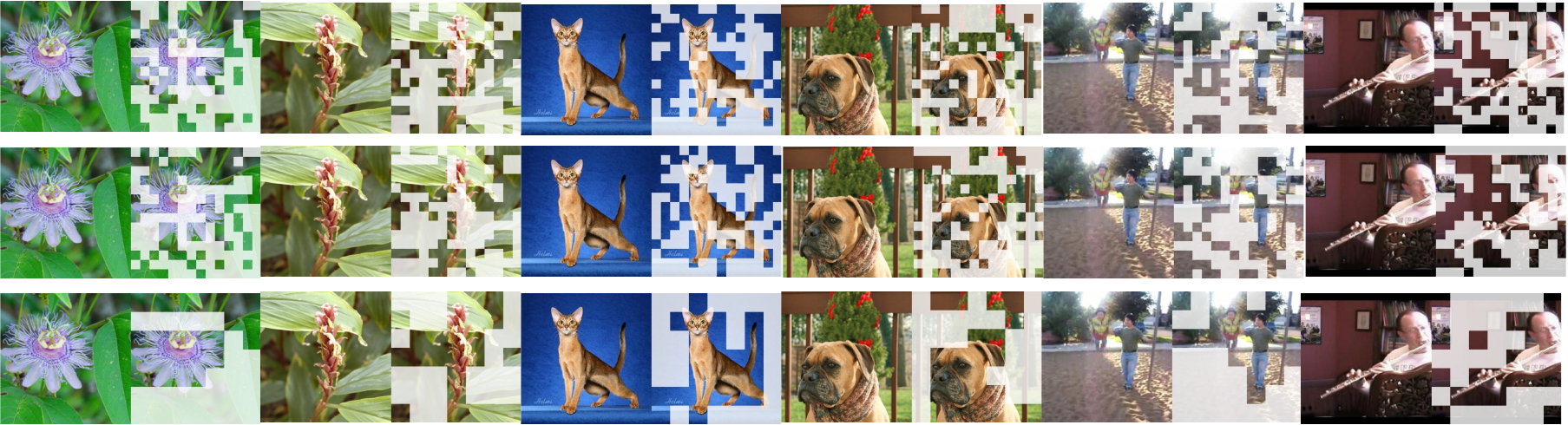}
  \end{center}
  \vspace{-1.4ex}
\caption{Visualization of Scoring Functions for Patch Token Pruning: The scoring functions for patch token pruning are visualized as follows: \textbf{Top row} -- Label-Driven Ranking Score, \textbf{middle row} -- Maximum Confidence Score, and \textbf{bottom row} -- Feature Preservation Score.}
    \label{fig:viz_scoring}
\end{figure*}

\section{\label{sec:experiments}Experiments}
In this section, we present our experimental results, demonstrating the effectiveness of our approach in improving CLIP's efficiency of inference. We first describe the experimental settings, followed by an investigation of the Golden Ranking. Subsequently, we present the results of prompt tuning, which integrates learnable tokens into the CLIP model to recover the performance loss resulting from token pruning. Finally, we conduct an ablation study to evaluate the effectiveness of our approach.

\subsection{Experimental Setting}
We conduct experiments using the ViT-B/16 model as the pre-trained CLIP's visual encoder on seven datasets: Caltech101~\cite{fei2004cal101}, OxfordPets~\cite{parkhi2012pets}, Flowers102~\cite{nilsback2008flower}, Food101~\cite{bossard2014food}, FGVCAircraf~\cite{maji2013aircraft}, DTD~\cite{cimpoi2014DTD},
UCF101~\cite{soomro2012ucf101} and
ImageNet~\cite{imagenet}
These datasets are chosen to represent a diverse range of image classification tasks, including object recognition, scene classification, and fine-grained classification. The official train/test splits were used for all datasets. While no further training data is used for zero-shot experiments (\ie, without prompt tuning), 16 images per class are used for prompt-tuning experiments, following the training splits of~\cite{Zhou2022CoOp}.

\subsection{\label{sec:golden_ranking_exp}Golden Ranking}
We begin by evaluating the effectiveness of the Golden Ranking in a Zero-Shot setting.
In this experiment, we present our investigation into the Golden Ranking. We focus on four aspects: (1) evaluating its effectiveness in classification performance; (2) examining the predictor's ability to accurately approximate the Golden Ranking; (3) assessing the classification performance using predictor-estimated rankings; and (4) evaluating the generalizability of the predictor across different datasets.

\paragraph{Effectiveness of Golden Ranking.}
\cref{sec:golden_ranking} introduces three distinct scoring functions to establish the utility of each token: \emph{Label-Driven Ranking Score}, \emph{Maximum Confidence Score}, and \emph{Feature Preservation Score}. 
Table~\ref{tbl:golden_ranking} (columns 3 to 7) shows the classification performance of the pruned CLIP model while utilizing the ground truth Golden Rankings at various pruning rates. Interestingly, pruning a significant portion of patch tokens using the label-driven ranking scores results in substantial accuracy improvements over the original 0-shot performance without pruning. This can be attributed to the ability of the label information to effectively identify and eliminate distractor patch tokens, which might mislead the CLIP model. As for the maximum confidence and feature preservation scores, the overall performance remains relatively stable even after significant pruning, but their effectiveness varies depending on the dataset.

\input{tables/cross_data}
\input{tables/runtime}
\input{tables/comparison}
\input{tables/main_exp}

\paragraph{Golden Ranking Predictability.}
Although the label-driven ranking score shows the highest accuracy in Table~\ref{tbl:golden_ranking}, ground truth rankings are not available during inference. Thus, high performance can only be achieved if the predictor can accurately estimate the Golden Ranking. To quantify this, we measured the percentage of the top 100 predicted tokens that match the top 100 tokens from the Golden Ranking.
As shown in~\cref{tbl:golden_ranking} (column 8), the label-driven and maximum confidence ranking scores are more challenging to predict accurately than the feature preservation score.

\begin{figure}[h]
  \begin{center}
    \includegraphics[width=0.38\textwidth]{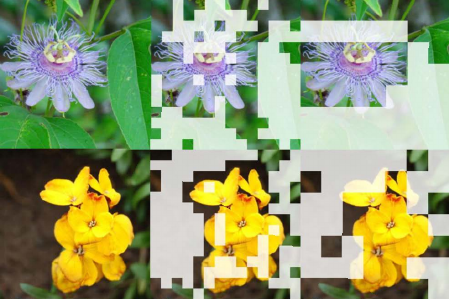}
  \end{center}
    \vspace{-1.4ex}
  \caption{This figure compares token pruning methods at the 50\% keep rate: the \textbf{middle column} shows CLS attention weight-based pruning, and the \textbf{right column} features our Feature Preservation Score method.}
  \label{fig:viz_CLS}
\end{figure}

For a deeper understanding of the predictor's performance, we visualized the tokens pruned by each of the scoring functions in \cref{fig:viz_scoring}. As can be seen, the tokens pruned according to the label-driven and maximum confidence scoring function seem to be less intuitive (from a human perspective) than those pruned by the feature preservation score. This suggests that the ``distractor'' tokens identified by the label-driven ranking score are likely to be caused by subtle visual patterns that are challenging for the predictor to learn. The feature preservation score, on the other hand, seems to be more effective in identifying redundant tokens, such as background elements, which are less likely to be crucial for classification.

\paragraph{Predictor-Based Pruning.}
To assess the predictor's performance, we conduct zero-shot inference utilizing the predictor for token pruning. As shown in Table~\ref{tbl:golden_ranking} (last 4 columns), a predictor trained to regress the feature preservation score generally achieves superior accuracy compared to the other two scoring functions across all keep rates. These findings suggest that the higher predictability of the feature preservation score makes it more suitable for token pruning in practice.

\paragraph{Comparison to CLS Attention Pruning.}
We also compared our pruned tokens against the more common CLS attention pruning. As shown in \cref{fig:viz_CLS}, the proposed scoring function seems to provide a more stable pruning set.
Since, unlike CLS attention, the predictor is trained to identify tokens that do not affect the output embeddings, crucial tokens for object identification are more likely to be preserved. As a result, predictor-based pruning is more effective in maintaining classification performance at lower keep rates, as shown in~\cref{fig:performance_rate_change} and expanded in Supplementary Material for a variety of datasets.

\paragraph{Generalizability of Predictor.}
Despite the encouraging results of~\ref{tbl:golden_ranking}, one potential drawback of the proposed approach is the fact that the predictor is directly trained on images of the target task. However, we found that the predictor generalizes well across datasets, as highlighted in \cref{tab:cross_dataset}.
Each predictor, despite being trained on a singular dataset (the columns in the table) to approximate the feature preservation score, can be applied to prune tokens from other datasets without significant drops in performance. Thus, despite requiring additional training data, these results show that our predictor does not require training data from the downstream task and thus retains CLIP's capability for open-dictionary recognition.

\paragraph{Runtime.}
To better assess computational efficiency, we measured the GFLOPs and inference time of the pruned CLIP model with a keep rate of 60\%. The results, presented in Table~\ref{tab:gflops_and_time}, show that our method significantly reduces the computational cost of existing V-L methods like CLIP, showing a decrease of over 30\% in GFLOPs and approximately 50\% in inference time.

\subsection{Comparison with Existing Works} Our work aims to enhance the efficiency of Vision-Language models like CLIP by introducing a lightweight token pruning predictor. This approach differs from traditional methods that utilize CLS attention weights, as it approximates the Golden Ranking, demonstrated in \cref{fig:performance_rate_change} to be more effective at maintaining classification performance with lower keep rates. We compare our method against existing token pruning techniques for ViT models, such as EViT~\cite{Liang2022EViTEV}, AViT~\cite{yin2022avit}, and ToMe~\cite{bolya2022tome}, which preserve CLIP's pre-trained weights without additional training. As shown in Table~\ref{tab:performance_comparison}, our method consistently outperforms these alternatives across various datasets. Furthermore, our assessment using both dataset-specific and agnostic predictors reveals that token pruning with a general predictor trained on ImageNet achieves comparable accuracy to dataset-specific predictors, confirming its effectiveness and versatility even in zero-shot settings.

\subsection{\label{sec:prompt_tuning_exp}Prompt Tuning}

\paragraph{Prompt Tuning Ablation.}
As discussed in \cref{sec:prompt_tuning}, regardless of the predictor accuracy in identifying redundant patches, the integration of learnable tokens is crucial for (1) enhancing the zero-shot prediction by better aligning visual and class embeddings and (2) recovering the performance loss due to token pruning. To evaluate the effectiveness of prompt tuning, we ablated the integration of text and visual prompts across all datasets.
The results, shown in \cref{tbl:main_exp}, support several conclusions. First, as shown in the first two rows, pruning without tuning results in a 1.6\% accuracy drop on average across all datasets. While acceptable, this drop can be mitigated with learnable tokens. Second, the integration of text prompts (T-tuning) significantly improves the performance of the baseline method (without pruning), from 67.9\% to 79.6\% on average. However, text prompts do not help mitigate the performance gap after pruning.
Finally, while visual prompts (V-tuning) do not help with baseline performance significantly, they are crucial for recovering the performance loss due to the pruning of visual tokens.

\noindent \textbf{Prompt Tuning with Varying Keep Rates.}
\cref{fig:performance_rate_change} compares the accuracy of token pruning, with and without prompt tuning for varying keep rates.~\cref{fig:performance_rate_change} also compares the proposed predictor-based tuning with a CLS attention-based method. As can be seen, the proposed method is more effective in maintaining classification performance at lower keep rates. This improved efficiency is particularly pronounced when learnable visual tokens are integrated into the model, showing nearly unchanged performance even after pruning as much as 50\% of the tokens on the Caltech101 dataset.

%% file: tables/cross_data.tex
\begin{table*}[t]
    \centering
    
    \resizebox{0.75\textwidth}{!}{%
    \begin{tabular}{
      >{\raggedright\arraybackslash}p{2.1cm}
      S[table-format=2.1]
      S[table-format=2.1]
      S[table-format=2.1]
      S[table-format=2.1]
      S[table-format=2.1]
      S[table-format=2.1]
      S[table-format=2.1]
    }
    \toprule
    \multicolumn{1}{c}{} & \multicolumn{7}{l}{\bf Predictor Training Dataset $\rightarrow$} \\
    {\bf Test Dataset $\downarrow$} & {Caltech101} & {OxfordPets} & {Flowers102} & {Food101} & {FGVCAircraft} & {DTD} & {UCF101} \\
    \midrule
    Caltech101 & 92.0 & 92.4 & 91.3 & 92.5 & 92.5 & 92.0 & 92.0 \\
    OxfordPets & 86.2 & 86.7 & 84.7 & 85.3 & 86.4 & 85.6 & 86.2 \\
    Flowers102 & 67.9 & 67.9 & 67.2 & 67.6 & 68.3 & 68.2 & 67.0 \\
    Food101 & 82.3 & 82.0 & 81.5 & 81.9 & 83.4 & 82.0 & 82.7 \\
    FGVCAircraft & 19.9 & 20.8 & 18.9 & 20.5 & 21.7 & 19.4 & 19.1 \\
    DTD & 45.3 & 44.3 & 44.7 & 43.9 & 44.0 & 44.3 & 45.1 \\
    UCF101 & 61.2 & 60.6 & 59.8 & 60.2 & 61.9 & 60.6 & 60.8 \\
    \bottomrule
    \end{tabular}%
    }
    \caption{\label{tab:cross_dataset}Cross-dataset generalizability of the golden ranking predictor. Columns indicate the predictor training dataset and rows the testing distribution. Models are evaluated through 0-shot recognition with a keep rate of 50\%.}
\end{table*}

%% file: tables/runtime.tex
\begin{table}[h]
        \centering
        \resizebox{\linewidth}{!}{
            \begin{tabular}{@{}lcc@{}}
            \toprule
            Methods         & GFLOPs & Time (ms) \\ 
            \midrule
            CLIP           & 23.4  & 3.5    \\
            $\hookrightarrow$ w/ pruning (60\% keep rate) & 16.8   & 1.6    \\
            \midrule
            CLIP+Prompt Tuning           & 27.6  & 4.0    \\
            $\hookrightarrow$ w/ pruning (60\% keep rate) & 20.9   & 2.0    \\ \bottomrule
            \end{tabular}
        }
        \caption{\label{tab:gflops_and_time}Computational efficiency in terms of GFLOPs and inference time per image. Measured on an NVIDIA A4500 GPU.}
\end{table}

%% file: tables/comparison.tex
\begin{table*}[ht]
    \centering
    \resizebox{0.85\linewidth}{!}{%
    \begin{tabular}{lcccccccc|c|c}
        \toprule
        \textbf{Method} & \textbf{Caltech101} & \textbf{OxfordPets} & \textbf{Flowers102} & \textbf{Food101} & \textbf{FGVCAircraft} & \textbf{DTD} & \textbf{UCF101} & \textbf{ImageNet} &\textbf{Avg} &\textbf{GFLOPs} \\ \hline
        EViT~\cite{Liang2022EViTEV} & 92.5& 87.1 & 67.1 & 80.3 & 23.3& 43.3 & 63.3 & 58.1 & 64.3 & 16.8\\
        A-ViT~\cite{yin2022avit} & 91.4& 83.2 & 67.7 & 82.3 & 21.7& 43.5 & 63.3 & 57.6 & 63.8 & 16.8\\
        ToMe~\cite{bolya2022tome} & 91.5& 87.2 & 67.7 & 82.4 & 20.4& 41.5 & 64.9 & 58.3 & 64.2 & 16.8\\
        \hline
        \rowcolor{tabhighlight}
        Ours & 93.4 & 88.1 & 67.9 & 82.8 & 22.9 & 43.7 & 65.2 & 59.5 &\textbf{65.4} & 16.8\\
        \rowcolor{tabhighlight}
        Ours-IN & 93.6 & 87.7 & 69.6 & 84.0 & 21.6 & 44.3 & 63.0 & 59.5 & \textbf{65.4} & 16.8\\
        \bottomrule
    \end{tabular}}
    \caption{\label{tab:performance_comparison}Comparison with prior token pruning methods. We apply these methods to CLIP's ViT. To ensure the same level of computational cost, we prune tokens at the 4th, 6th, 8th, and 10th layers of the CLIP's ViT, eliminating 20 tokens at each specified layer. For fair comparison to prior work, we evaluate our token pruning without further prompt tuning (Phase III). \textbf{Ours-IN} refers to the predictor trained solely on ImageNet and then applied to all eight datasets.} 
\end{table*}

%% file: tables/main_exp.tex
\setlength{\tabcolsep}{3.3pt}
\begin{table*}[t]
\centering
\tiny
\newcommand{\rotbox}[1]{\rotatebox{40}{#1}}
 \resizebox{0.95\textwidth}{!}{
\begin{tabular}{ccc|ccccccc|c|c}
\toprule
\bf T-tuning & \bf V-tuning & \bf Pruning& \bf Caltech101 &\bf OxfordPets &\bf Flowers102 & \bf Food101 &\bf FGVCAircraft &\bf DTD &\bf UCF101 &\bf Average &\bf GFLOPs\\ 
\midrule
&&&93.5 &89.5 &70.5 &86.0 &23.4 &45.1 &67.0&67.9&23.4\\
&&\cmark&93.4 &88.1 &67.9 &82.8 &22.9 &43.7 &65.2&66.3&16.8\\
\midrule
\cmark&&&95.2 &92.5 &95.7 &84.8 &37.3 &70.0 &81.7 &79.6&26.9  \\
\cmark&&\cmark&94.5 &91.7 &93.8 &82.3 &34.7 &67.4 &81.0 &77.9&18.9\\
\midrule
\cmark&\cmark&&95.6 &92.3 &95.5 &84.9 &37.6 &69.7 &82.7 &79.8&27.6 \\
\cmark&\cmark&\cmark&95.1 &92.2 &95.4 &84.2 &38.8 &68.9 &81.8 &79.5&20.9\\
\bottomrule
\end{tabular}
}
\caption{Accuracy evaluation with learnable tokens. Results from applying prompt tuning to CLIP's ViT to recover accuracy loss from pruning 40\% of patch tokens. 'T-tuning' uses 16 text prompts, and 'V-tuning' uses 16 visual prompts, both shared across dual encoders. Pruning occurs progressively at the 4th, 6th, 8th, and 10th layers, removing 20 tokens each.}
\label{tbl:main_exp}
\end{table*}

%% file: content/5-conclusion.tex
\section{Conclusion}
\label{sec:conclusion}

In this work, we introduce a novel framework designed for pruning patch tokens in CLIP's ViT, effectively addressing the computational intensity typically associated with these models. At the heart of our approach is the ``Golden Ranking'' concept, which methodically ranks patch tokens based on scoring functions. A key innovation in our method is the deployment of a lightweight predictor, trained to closely approximate this Golden Ranking. Furthermore, to mitigate the inevitable performance loss resulting from the pruning process, we integrate learnable text and visual tokens into our framework. These tokens, especially visual tokens, were shown to play a pivotal role in compensating for potential performance degradation, ensuring the model's output remains accurate post-pruning. Our extensive experiments across a variety of datasets have demonstrated that our framework can achieve a substantial reduction in patch tokens, by up to 40\% in CLIP's ViT, while maintaining comparable performance (only 0.3\% lower accuracy). %

%% file: content/6-acknowledgement.tex
\section*{Acknowledgement}
\noindent This work was partially supported by the National Science Foundation under Grant No. 2006394.

%% file: content/7-supplementary.tex
\begin{figure*}[t]
  \begin{center}
    \includegraphics[width=0.80\textwidth]{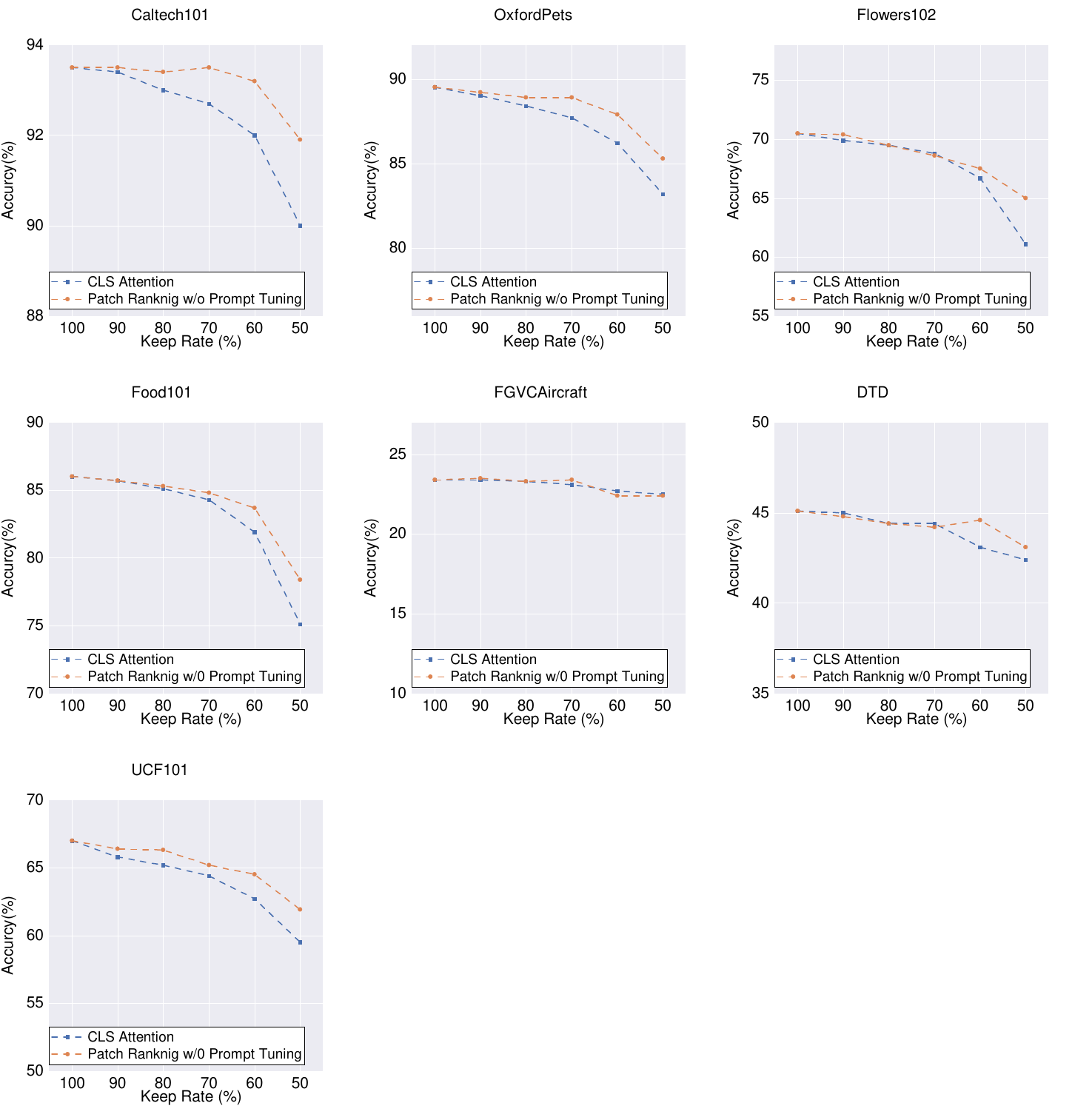}
  \end{center}
  \caption{This figure compares the classification accuracy between the CLS attention method and our Patch Ranking approach, both without fine-tuning post-token pruning. CLS attention employs CLS attention weights to rank tokens, whereas Patch Ranking utilizes the Feature Preservation Score for this purpose. Token removal occurs at the first layer of CLIP's ViT. We present classification accuracy across different keep rates, ranging from 100\% to 50\%, highlighting the differential impact of each method on model performance as the number of pruned tokens increases.}
  \label{fig:cls_vs_ours_full}
\end{figure*}

\section{Additional Results Comparison}
\paragraph{Comparison with CLS Attention}
In prior works, using CLS attention weight to rank the importance of patch tokens has been a prevalent method for enhancing the efficiency of Vision Transformers (ViTs). However, this approach is less effective for CLIP's ViT due to its dual-modality structure. Addressing this limitation, we introduce 'Patch Rank,' a novel framework tailored for CLIP's ViT. To assess the efficacy of Patch Rank, we conduct a comparative analysis with the CLS attention method across seven datasets, evaluating performance at keep rates ranging from 100\% to 50\%. Token pruning was executed at the first layer of CLIP's ViT to optimize computational savings. Importantly, neither method performs fine-tuning after token pruning. As shown in Figure~\ref{fig:cls_vs_ours_full}, our Patch Ranking consistently demonstrates higher accuracy than CLS attention across all keep rates and datasets. Notably, our method shows a significant advantage over CLS attention, especially at lower keep rates (60\% and 50\%). This outcome indicates the ability of Patch Rank to precisely eliminate less informative patch tokens while minimizing the loss in accuracy, thereby affirming its effectiveness in the nature of CLIP's ViT.
\section{Ablation study}
\paragraph{Architecture of Predictor}
To construct our predictor, we selected three different architectures: (1) MLP, which consists of a 256-dimensional hidden layer, layer normalization, GLUE, and a 196-dimensional hidden layer; (2) Transformer, specifically a Transformer-encoder block; and (3) Mix-MLP, which is a single block configuration. To assess the performance of these architectures, we evaluated their top-100 matching rates and pruning effectiveness across various keep rates, from 80\% to 50\%. As depicted in Table~\ref{tbl:predictor}, Mix-MLP emerges as the most effective, achieving the highest matching rate. Regarding the performance in token pruning, Mix-MLP demonstrates stable results across all datasets, and notably, it significantly outperforms the other architectures in the UCF101 dataset. This superiority of Mix-MLP can be attributed to its optimal capacity for learning and applying the Golden Ranking, coupled with its ability to avoid overfitting the training dataset.
\input{tables/predictor}
\paragraph{Token Pruning Locations}
\input{tables/prune_location}
In our exploration of token pruning locations within CLIP's Vision Transformer, we conducted an in-depth analysis to determine the impact of varying pruning depths on model performance. This involved progressively pruning an equal number of patch tokens at different layers while maintaining a consistent keep rate of 60\%. The results are shown in Table~\ref{tbl:prune_location}. It focuses on four distinct combinations of pruning locations, ranging from shallower to deeper layers within the network. Despite a slight margin favoring pruning patch tokens at deeper layers, the overall average performance across all datasets remains notably consistent. This suggests that our predictor can adapt to different layers within the network, accurately estimating rankings, and identifying redundant tokens across various depths. Specifically, the minimal variation in performance across different pruning configurations indicates that our approach maintains the predictor's ability regardless of the specific layers targeted for token reduction.

%% file: tables/predictor.tex
\setlength{\tabcolsep}{5.0pt}
\begin{table}[h]
\small
\centering
\resizebox{1.0\linewidth}{!}{
\begin{tabular}{ll|cccccc}
\toprule
\multirow{2}{*}{\bf Dataset} &
\multirow{2}{*}{\begin{tabular}[c]{@{}c@{}} \bf Arch. \end{tabular}}
&\multirow{2}{*}{\bf Matching rate}&\multicolumn{5}{c}{\bf Predictor}  \\
\multicolumn{2}{l|}{}&                             
& 100 & 80 & 70 & 60 & 50  \\ \hline
\multirow{3}{*}{Caltech101} 
& MLP &76.5&93.5&93.3&93.2&92.7&91.0\\ 
& Trans. &73.4&93.5&93.4&\bf93.3&92.8&\bf91.2 \\ 
& Mix-MLP  &\bf78.0&93.5&\bf93.6&93.2&\bf93.4&91.0\\ 
\hline
\multirow{3}{*}{OxfordPets} 
& MLP &75.7&89.5&\bf89.2&88.5&88.0&84.5\\ 
& Trans. &72.9&89.5&89.0&\bf89.0&88.0&\bf85.5 \\ 
& Mix-MLP  &\bf78.2&89.5&88.6 &88.4 &\bf88.1 &83.5 \\ 
\hline
\multirow{3}{*}{Flowers102} 
& MLP &69.2&70.5&69.5&\bf69.2&67.4&\bf64.6 \\ 
& Trans. &56.2&70.5&\bf69.8&69.0&67.5&60.9  \\ 
& Mix-MLP  &\bf71.9&70.5&69.6 &68.8 &\bf67.9 &63.9\\ 
\hline
\multirow{3}{*}{Food101} 
& MLP &70.4&86.0&85.5&84.8&\bf83.7&78.3 \\ 
& Trans. &69.7&86.0&\bf85.7&\bf85.0&83.8&\bf78.5  \\ 
& Mix-MLP  &\bf71.9&86.0&85.1 &84.2 &82.8 &75.8\\ 
\hline
\multirow{3}{*}{FGVCAircraft} 
& MLP &85.2&23.4&23.5&23.1&\bf23.1&\bf22.8 \\ 
& Trans. &84.6&23.4&\bf23.6&\bf23.6&22.9&\bf22.8  \\ 
& Mix-MLP  &\bf89.3&23.4&23.2&23.2 &22.9 &22.4\\ 
\hline
\multirow{3}{*}{DTD} 
& MLP &\bf77.5&45.1&44.6&44.3&43.7&41.9 \\ 
& Trans. &62.1&45.1&44.5&\bf44.9&\bf43.8&42.1  \\ 
& Mix-MLP  &72.3&45.1&\bf45.0&44.2&43.7&\bf43.0\\ 
\hline
\multirow{3}{*}{UCF101} 
& MLP &77.5&67.0&61.8&58.2&53.6&43.2 \\ 
& Trans. &51.2&67.0&62.2&60.0&53.5&43.5  \\ 
& Mix-MLP  &\bf79.7&67.0&\bf66.6 &\bf65.9 &\bf65.2 &\bf60.9\\ 
\bottomrule
\end{tabular}
}
\vspace{10pt}
\caption{Design Choices for the Predictor: This table explores three different architectures employed as predictors: Multilayer Perceptron (MLP), Transformer-encoder block (Trans.), and Mix-MLP. We evaluate these architectures based on their top-100 matching rates and classification accuracy across various keep rates, ranging from 100\% to 50\%. Token pruning is executed at the 4th layer of CLIP's ViT, aiming to assess the effectiveness of each architecture in maintaining accuracy while managing token redundancy.}
\label{tbl:predictor}
\end{table}

%% file: tables/prune_location.tex
\begin{table*}[t]
\centering
\small
\newcommand{\rotbox}[1]{\rotatebox{0}{#1}}
\begin{tabular}{c|ccccccccc}
\toprule
 \bf Pruning Locations & \bf \rotbox {Caltech101} &\bf \rotbox{OxfordPets} &\bf \rotbox{Flowers102} &\bf \rotbox{Food101} &\bf \rotbox{FGVCAircraft} &\bf \rotbox{DTD} &\bf \rotbox{UCF101} &\bf Avg. \\ \hline
2, 3, 4, 5&94.4&92.3&94.3&82.0&37.9&67.6&81.7&78.6 \\
4, 5, 6, 7&94.3&92.1&95.3&82.2&39.5&68.1&82.0&79.1 \\
1, 3, 5, 7&94.8&91.6&94.4&82.0&39.0&67.8&81.8&78.9\\
4, 6, 8, 10&95.3&91.4&94.5&83.0&40.0&68.4&83.0&\bf79.2   \\

\bottomrule
\end{tabular}
\caption{Performance analysis across different pruning locations: In this experiment, we maintained a keep rate of 60\% and progressively pruned equal quantities of patch tokens at four distinct layers within CLIP's ViT. We examined four different combinations of pruning locations to evaluate how varying the pruning layers within the network layers affects overall model performance.}
\label{tbl:prune_location}
\end{table*}